# EXAMPLES AS INTERACTION: ON HUMANS TEACHING A COMPUTER TO PLAY A GAME

DIMITRIS KALLES

*Hellenic Open University, Tsamadou 13-15,*
*Patras, 26222, Greece*
*kalles@eap.gr*

ILIAS FYKOURAS

*Hellenic Open University, Tsamadou 13-15,*
*Patras, 26222, Greece*
*fykouras@sch.gr*

This paper reviews an experiment in human-computer interaction, where interaction takes place when humans attempt to teach a computer to play a strategy board game. We show that while individually learned models can be shown to improve the playing performance of the computer, their straightforward composition results in diluting what was earlier learned. This observation suggests that interaction cannot be easily distributed when one hopes to harness multiple human experts to develop a quality computer player. This is related to similar approaches in robot task learning and to classic approaches to human learning and reinforces the need to develop tools that facilitate the mix of human-based tuition and computer self-learning.

## 1. Introduction

Programming a computer what to do is a task that lies at one end of the spectrum spanning interactive behavior of computers. Shifting a bit, using a shell-like development environment, that completes commands as soon as there are no alternatives to what the user will type, is surely a behavior that is worth being shifted a little from that far-end point of nearly-no-interaction. About two decades ago that behavior would likely have also been called "intelligent" even though it might have just employed a trie for indexing the shell commands. Today, of course, we reserve the "intelligent" tag for computing behaviors that are far richer in content and structure than the automatic completion of unfinished commands.

Extending the programming parable we note that today we have a range of tools at our disposal that are built with deep knowledge of the development process and intelligent compilers that generate speculative code; all these suggest that software development now demonstrates some noteworthy richness in its human-computer component. Actually, such shifts along the capabilities of the "interaction spectrum" also highlight how machine learning has also evolved over the last two-three decades. While earlier we addressed applications of learning that only used propositional data, we now try to infer structure, to deal with temporal aspects of learning problems and to use intelligent techniques not merely as classifiers but as discoverers of new knowledge.

It is thus only natural that some machine learning techniques have made inroads into problems that demonstrate a significant human-computer interaction component. Games are a prime example of such an application; Shannon[1] and Samuel[2] provided the first stimulating examples, Deep Blue defeated Kasparov at chess in 1997[3] and, more recently, Schaeffer's team solved checkers completely.[4] Of course, in-between those years and quite before the successes of computer programs, there has been a thriving market for strategy games playing machines for the public. The advances of machine learning techniques now allow us to tackle a much more difficult and challenging question: how can a person instruct a computer to play a game by simply showing it how to play? This is a new human computer interaction problem – we are given a computer that is equipped with a generic learning mechanism and we have the task to present to it a "syllabus" of experience that will allow it to formulate playing knowledge.

This paper is about using humans as experts who attempt to teach a computer how to play a strategy board game. We pursue this line of work because we are both interested in exploring the distinct human styles in instructing machines and, on top of that, to investigate whether expert playing behavior can be generalized from brief, not-so-expert, training sessions. Our main experimental result is that while individual training sessions between a human and a computer can improve the computer's performance, a straightforward composition of individually learned behaviors is not yet possible. To arrive at this observation we designed and carried out about 1,000 human-vs.-computer games and about 500,000 computer-vs.-computer games.

For our work, we used two distinct groups of users, one consisting of high-school students and one consisting of their tutors, across various disciplines. The workbench we use is a relatively simple strategy board game that, for legacy reasons, is called *RLGame*, since it uses reinforcement learning as the sole learning mechanism to infer how to play by observing games in action.

The reason we are interested in investigating both pupils and teachers is primarily because these two groups have such a huge gap in their academic development, experience and perception of how one instructs, that they constitute a fascinating user group for testing artificial intelligence based interaction on the assumption that a computer starts by knowing nothing.

The rest of this paper is structured in five sections. We first review the work that is most closely related to our research, on aspects of human-computer interaction in machine leaning in games. We then review our work on the subject, present the rules of the games we use as a workbench and the experiments conducted so far, to draw the base upon which we extend our work. We then present the experiments we designed and carried out with high school pupils and their teachers, as well as the automated experiments that help us make a comparison. We then discuss our findings and we conclude, also presenting the ramifications of our work for human-computer interaction in learning systems at large.

## 2. A brief review of related work

Thomaz and Breazeal[5] offer a concise yet informative account of recent attempts to exploit human interaction in human-trainable systems. Their summary review identifies at least three major dimensions of such systems that can be viewed as control knobs in the human-computer interaction loop.

One first such dimension has to do with the classic (and, quite mainstream) approach to personalization and adaptability of user interfaces where one simply collects raw items of user behavior and then attempts to elicit structure, intention or habit from such data.[6,7] The human may be aware of the data collection process but no specific attempt is made to act as "teacher". This approach, of course, has its roots to one archetypal design peculiarity of knowledge-based systems: the elicitation of knowledge by various techniques except interviews, by observation or by measurement, to name a couple. Any result in investigating how to best utilise expert involvement will likely have significant ramifications in the design of such systems.

A further related dimension is the identification of how control is exerted. While the first dimension clearly associates learning with a passive computer, one might elect a more proactive approach whereby the computer (as a student) identifies gaps in its knowledge and generates actions (or, queries) to fill these gaps; this can be done fully or partially (for example, by generating actions that minimize some measure of the knowledge gap[8]). But, learning in games is a domain where such identification of where control is exerted is not straightforward, since a system may generate the right queries but the human tutor may respond unpredictably and not according to the plan.

Yet another dimension is whether the human tutor is a part of the training mechanism at all times; this would mean that every step is subject to human critique. This approach draws on nature inspired paradigms, such as clicker-training[*], and its representatives are systems where continuous but simple feedback suffices to train robots.[9,10,11]

An approach that has attracted a lot of attention is to use the human tutor for focused guidance and, subsequently, to use some mechanism of self-improvement up to a point where, again, the human tutor will be called into action to provide a correction of direction, if required; this can go *ad infinitum*.

This approach draws heavily on a human learning analogy. When confronted with strategy issues in an unknown field humans usually learn by trial and error. By confronting problems in a new domain we slowly develop measures of success and of measuring up the difficulty of the problem at hand. But, when on one's own, the selection of problems is a delicate issue and can easily lead to deadlocks and a feeling of underachievement. A problem that is addressed at too fine a resolution focuses us too much and we may be unable to generalize if we are novices. On the other hand, if the resolution is not fine enough, we will be unable learn something specific upon which to build on.

---

[*] A clicker emits a brief and sharp sound upon pressing it. This sound does not mean anything by itself for the animal but a trainer can associate it with things that the animal instinctively finds rewarding such as food. After a certain number of such associations, the clicker starts meaning that a reward will come soon; this is the point where the trainer can start guiding the animal towards the desired behavior.[9]

Much more effective is the employment of a tutor, if we can afford one. Our research navigates on the fine line between the sparseness and density of learning examples when the computer serves as the student[12,13,14] and the goal is to establish some examples of successfully tuning the "syllabus". Along that direction we expect that the length and the content of a training session will slowly become evident (but, automatically so) if we spend enough time even with very simple feedback (and, here, we draw again a contrast to conventional clicker training[9], where the length of the training sessions is fine tuned by trainers).

A considerable line of research involves the merging of low-level associative (similarity-based) search with higher-level (for example, spatial awareness) cognitive-based rules.[15] We believe that this resonates well with approaches which emphasize an interactive behaviour in learning[5,16,17,18], where one would ideally switch from focused human training to autonomous crawling between promising alternatives. Our own results confirm this approach and are in line with observations that there is research value in investigating human impact in learning a new game precisely because it is worth exploring how a *disturbance* affects learning.[†]

### 3. A brief background on a strategy game workbench

The game[19] is played by two players on an $n \times n$ square board. Two $a \times a$ square bases are on opposite board corners; the white player starts off the lower left base and the black player starts off the upper right one. Initially, each player possesses $β$ pawns. The goal is to move a pawn into the opponent's base. Currently, we use $n = 8$, $a = 2$ and $β = 10$.

The base is considered as a single square, therefore a pawn can move out of the base to any adjacent free square. Players take turns and pawns move one at a time. A pawn can move vertically or horizontally to an adjacent free square, provided that the maximum distance from its base is not decreased (so, backward moves are not allowed). The distance from the base is defined as the maximum of the horizontal and the vertical distance from the base. A pawn that cannot move is lost (more than one pawn may be lost in one move). A player also loses by running out of pawns.

---

[†] We use the term *disturbance* to denote the surprises that a human tutor can present to a learning machine, such as an unanticipated move due to human long-term planning or a skewed reward that may be due to a specific mental or sentimental situation.

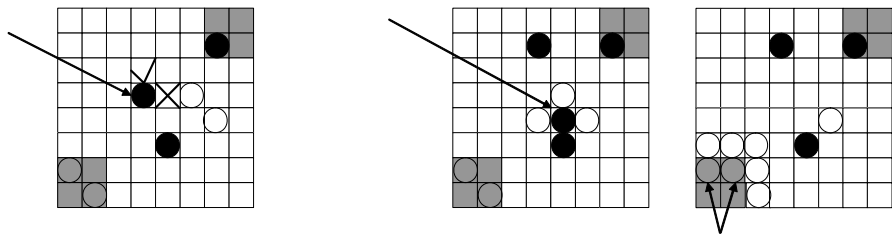

Fig. 1. Examples of game rules application.

The leftmost board in Fig. 1 demonstrates a legal and an illegal move for the pawn pointed to by the arrow, the illegal move being due to the rule that does not allow decreasing the distance from the home (black) base. The rightmost boards demonstrate the loss of pawns, with arrows showing pawn casualties. A "trapped" pawn automatically draws away from the game; so, when there is no free square next to the base, the rest of the pawns of the base are lost.

In *RLGame*[‡] the *a priori* knowledge of the system consists of the rules only. To judge what the next move should be, we use reinforcement learning[20] to learn an optimal policy that will maximize the expected sum of rewards in a specific time, determining which action should be taken next given the current state of the environment. We approximate the value function on the game state space with neural networks[21], where each next possible move and the current board configuration are fed as input and the network outputs a score that represents the expectation to win by making that move[19], as initially adopted in Neurogammon.[22]

We have eventually a commonly used $\varepsilon$-greedy policy with $\varepsilon=0.9$ (the system chooses the best-valued action with a probability of 0.9 and a random action with a probability of 0.1), assigned to all states but the final the same initial value and updated values after each move through *TD*(0.5), thus halving the influence of credits and penalties for every backward step that we consider. For each neural network, the input layer nodes are the board positions for the next possible move plus a binary attribute on whether a pawn has entered an enemy base and plus some more binary attributes on whether the number of pawns in the home base has exceeded some thresholds, totalling $n^2-2a^2+10$ input nodes. The hidden layer consists of half as many hidden nodes. There is one output node; it stores the probability of winning when one starts from a specific game-board configuration and then makes a specific move.

Note that, while an optimal deterministic optimal policy does exist[23] for *RLGame*, using the neural networks for approximating the state space rules out learning of the optimal policy. We have not investigated if the initial, full state-space representation would make converging to the optimal policy computationally tractable.[24]

---

[‡] For legacy reasons, the game is still called *RLGame* to emphasize the technology used for learning.

### 3.1. *Reviewing the effects of expert involvement*

The initial experiments demonstrated that, when trained with self-playing, both players would converge to having nearly equal chances to win[19], and that self-playing would achieve weaker performance compared to a computer playing against a human player, even with limited human involvement.[16]

The next round of experimentations delivered support for the reward scheme used to-date[17] (see Table 1) and offered evidence that human-vs.-computer sessions of 10 games each ($HC^{10}$) would be preferable to $HC^1$ sessions, when interleaved with computer-vs.-computer sessions of 1,000 games each ($CC^{1000}$).[18]

Table 1. Rewards.

| Type of Event | Reward |
|---|---|
| Win | 100 |
| Loss | -100 |
| Any change in pawn number | Pawn difference scaled in [-100,100] for each player |

Along that session, we also devised a way to measure the relative effectiveness of the policies learned by two distinct approaches.[18] Assuming that we have available a player, *X*, with its associated white and black components, $W_X$ and $B_X$ (the components being the neural networks), we compare it to *Y* by first pairing $W_X$ with $B_Y$ for a $CC^{1000}$ session, then pairing $W_Y$ with $B_X$ for a further $CC^{1000}$ session, and subsequently calculating the number of games won and the average number of moves per game won (see Table 2 for an example showing how player *X*'s components collectively win *Y*'s).

Table 2. Comparative evaluation of learned policies *X* and *Y*.

|  | Games Won | | Average # of Moves | |
|---|---|---|---|---|
|  | White | Black | White | Black |
| $W_X$ vs. $B_Y$ | $715_X$ | $285_Y$ | 291 | 397 |
| $W_Y$ vs. $B_X$ | $530_Y$ | $470_X$ | 445 | 314 |

We then used the above reporting scheme in round-robin tournaments of learning policies (each player competing against every other player; note the difference with *elimination* tournaments where only winners advance to the next round) and observed that, in general, a low average number of moves per session was associated with one of the sides being a comprehensive winner as reported in the number of games won.

A novel type of expert was then used in the next experimental round, by employing a min-max algorithm for the moves of the white player. The key observation therein was the *pendulum* effect: the min-max tutor for the white player actually trains the black one by forcing it to lose; when the tutor goes, the black player overwhelms the white one, which has to adapt itself due to black pressure as fast as possible.[25]

## 4. Experimentation and analysis

For the work reported in this paper, the experimental session consisted of two distinct stages. During the first stage we collect data based on HC$^{40}$ sessions; this is the stage where humans do their best to teach a computer within a limited number (40) of games. During the second stage, the learned policies are paired in CC$^{1000}$ rounds of various types of elimination tournaments to obtain insight as to whether some individual attained a clearly good training of its "computer" players and to examine whether the composition of players may deliver a better player simply by self-play.

### 4.1. *Data collection via human-vs.-computer experimentation*

This session took part in a high school setting where one of the authors serves as teacher. Twenty students aged about 13 were assigned to play *RLGame* for 40 consecutive games each; the neural networks were initialized before the first game and were being updated throughout the HC$^{40}$ session. Additionally, twelve teachers were assigned to play *RLGame* for 40 consecutive games. The age, sex and specialty data of teachers is shown in Table 3$^\S$.

Table 3. Aggregate teacher data.

| Type | Data |
|---|---|
| Age | 25, 32, 45, 46$_4$, 47$_2$, 48, 50, 55 |
| Sex | 6 male, 6 female |
| Specialty | Languages$_2$, Math$_3$, Physics, French$_2$, Computing, Technology, Art, German |

Based on the current geometric configuration of *RLGame*, a player needs a minimum of 10 moves to navigate to the enemy base. Table 4 shows the average number of moves per player and player type, in increasing order.

Table 4. Average number of moves per player.

| | Average number of moves |
|---|---|
| Pupil | 10.1, 10.3, 10.5, 10.7, 10.8, 10.9$_3$, 11.1$_2$, 11.2$_2$, 11.6, 11.8, 11.9$_2$, 12.2, 13.9, 16.7, 23.0 |
| Teacher | 10.1, 10.3, 10.5, 10.6, 10.7, 11.5$_2$, 11.6, 12.0, 12.3, 12.7, 20.0 |

The data collection exercise had to be carefully planned. Both groups attended a short presentation on the rules of the games; furthermore, a one-page brochure that served as quick reference outlining the rules and the concept of the experimentation was distributed as an aid throughout the experiments. This brochure helped reinforce their capacity to look for information on their own and helped develop a positive attitude to the experiment, as the thoughtful preparation steps were noticed.

We instructed all users that we were asking them to attain two main goals, to win the computer, and to teach it. We emphasized "winning" to limit the degrees of freedom of our experiment and to ensure an as equal as possible footing of all learned policies. We also emphasized that the computer learns by wins, losses and pawn eliminations.

---

$^\S$ Subscripts in table values refer to the number of cases with that value, unless otherwise evident or defined.

However, there were aspects of the experimentation where we had to treat the pupils and their teachers completely separately:

- Pupils were enthusiastic about the prospect of participating in an experiment, since IT literacy was not a concern and the experiment took place in a familiar class context. Though the graphics of *RLGame* are not particularly attractive and though game play is not yet challenging for humans, being told that "the computers learns by what you teach it" was enough of a motivation to sustain the pupils' focus during the experimentation. Catering to the students' questions during game-play was a relatively straightforward and not demanding task.
- Teachers were less easy to co-ordinate, due to class scheduling, so experiments were carried out one-person-at-a-time with the presence of the resident author. It was instrumental to offer ample support for the teachers' reservations, which were only secondarily alleviated by pointing out to the scientific orientation of the experiment, the main concern being whether the experiment was a hidden assessment of their tutoring skills. Still, some of them refused to participate. We stress, referring the reader to Table 3, that we only report and use aggregate data and do not link individual sessions back to the people who carried them out.

For the sake of completeness we report that the pupils took between 2'14'' and 2'43'' to complete each game, whereas teachers took between 1'31'' and 2'53''.

### 4.2. *Data analysis via computer-vs.-computer experimentation*

In all CC experimentations that follow, the initial data consist by the pupils' and teachers' learned policies, represented by the neural networks (white and black) as trained during their separate $HC^{40}$ sessions.

#### 4.2.1. *Pupils-only tournaments*

The first tournament featured in this section paired students in successive $CC^{1000}$ elimination rounds but, at each subsequent round, the winning student advanced and re-used its initially available neural network.[**]

We remind the reader that even one move in any game is liable to trigger a change in the neural networks of both white and black players; so, while this can happen during the $CC^{1000}$ round, we do not use the modified neural networks for the next round. In effect, we have a *memory-less* elimination tournament.

Since the number of players in not a power of 2, some pupil players entered the elimination tournament after the first round. For a sample result, we present in Table 5 the final round of the tournament. We note that there is a noticeable similarity in how the black players outperform the white ones, both in term of games won and average number of moves; moreover the final round does not demonstrate a comprehensive difference between the two finalists (for clarity, we underline whatever is related to the winner). Going back to individual results from the elimination rounds (which we omit here for space economy) we observed that the further from the final we looked back, the easier both finalists seemed to advance to the next round. In effect, the elimination tournament indeed identified a (relatively-speaking) initially well trained player.

---

[**] To be succinct, we use the term "student" to refer to the student model as represented by the two neural networks for the white and black player. Furthermore, when pitting two models against each other, we actually need 2 distinct $CC^{1000}$ sessions (as shown in Table 2).

Table 5. The final round of the pupils' memory-less tournament (finalists: $P_{13}$, $P_{21}$).

|  | Games Won | | Average # of Moves | |
| --- | --- | --- | --- | --- |
|  | White | Black | White | Black |
| $W_{P\text{-}13}$ vs. $B_{P\text{-}21}$ | 260 | <u>740</u> | 358 | <u>140</u> |
| $W_{P\text{-}21}$ vs. $B_{P\text{-}13}$ | <u>376</u> | 624 | <u>375</u> | 190 |

We then experimented with a *synthesis* elimination tournament; therein winners advanced to the next round by using the modified neural networks, as evolved during the matches with their opponents. We review in Table 6 the final round of the tournament (the accents in the players indicate where they originated and how many elimination rounds they survived; fro example $P_7'''$ refers to how the $P_7$ evolved after three successful elimination rounds). We note that the $P_{17}'''$ finalist is a comprehensive winner, but going back to earlier games did not uncover a clear pattern of increasing ease as one moves towards the start of the tournament. This is an indication that a synthesis tournament indeed mixes the learned policies.

Table 6. The final round of the pupils' synthesis tournament (finalists: $P_7'''$, $P_{17}'''$).

|  | Games Won | | Average # of Moves | |
| --- | --- | --- | --- | --- |
|  | White | Black | White | Black |
| $W_{P\text{-}7'''}$ vs. $B_{P\text{-}17'''}$ | 82 | <u>918</u> | 761 | <u>107</u> |
| $W_{P\text{-}17'''}$ vs. $B_{P\text{-}7'''}$ | <u>970</u> | 30 | <u>15</u> | 92 |

Having performed the synthesis tournament we pitted $P_{sum}$ (which is $P_{17}'''$) against each original (pupil) player. $P_{sum}$ won 12 games (10 of which comprehensively) and lost the remaining 8 of them (all of which comprehensively).

### 4.2.2. *Teachers-only tournaments*

Table 7 presents the final round of the teachers' memory-less elimination tournament. However, this time, we do not observe the similarity seen in Table 5; this also relates to the fact that $T_{10}$ is a comfortable winner, even in the final. Going back to individual results from the elimination rounds we observed that $T_1$ advanced from the first round to the final through rather close wins, while $T_{10}$ had a close win in its first match and then advanced comfortably. This reiterates the observation we made in the corresponding pupils' case, that an initially well trained player won the tournament.

Table 7. The final round of the teachers' memory-less tournament (finalists: $T_1$, $T_{10}$).

|  | Games Won | | Average # of Moves | |
| --- | --- | --- | --- | --- |
|  | White | Black | White | Black |
| $W_{T-1}$ vs. $B_{T-10}$ | 422 | <u>578</u> | 280 | <u>170</u> |
| $W_{T-10}$ vs. $B_{T-1}$ | <u>785</u> | 215 | <u>10</u> | 58 |

Table 8 reviews the final round of the teachers' synthesis tournament. We note that $T_{12\cdots}$ is a comprehensive winner, but due to the short length of this tournament, we did not attempt to elaborate on how easy it might be to win at the start of the tournament.

Table 8. The final round of the teachers' synthesis tournament (finalists: $T_{3''}$, $T_{12\cdots}$).

|  | Games Won | | Average # of Moves | |
| --- | --- | --- | --- | --- |
|  | White | Black | White | Black |
| $W_{T-3''}$ vs. $B_{T-12\cdots}$ | 315 | <u>685</u> | 438 | <u>125</u> |
| $W_{T-12\cdots}$ vs. $B_{T-3''}$ | <u>642</u> | 358 | <u>112</u> | 207 |

After the teachers' synthesis tournament, however, we were surprised to see that $T_{sum}$ lost against each original (teacher) player and was able to put up a close performance to only 2 out of the 12 opponents.

### 4.2.3. *Combination matches*

A reasonable question arises from the above experiments: since the synthesis-produced players are not better than the individual players, does that mean that knowledge of how to play the game cannot be really synthesized but should be accumulated, game by game upon an initial *tabula rasa* configuration? Alternatively, one could question whether the individual players (20 + 12) did not offer any help to the training of the computer, after all. Still another explanation might be that the sessions were not long enough; of course since we have collected data only on $HC^{40}$ experiments we can test this hypothesis only on the subsequent CC sessions.

We focused on the $T_{sum}$ player, which produced the worst results. A straightforward extension was to evolve it for a further $CC^{10000}$ session; call the new player $T'_{sum}$. We then carried out a $CC^{1000}$ round-robin tournament between $T_{sum}$, $T'_{sum}$ and $P_{sum}$ and report the results of all three matches in Table 9, Table 10 and Table 11 (we drop the $_{sum}$ index in the tables). It is most interesting that while $P_{sum}$ overwhelms $T_{sum}$ (Table 10), all it takes to reverse this situation is to allow $T_{sum}$ to evolve to $T'_{sum}$. This is an indication that the $CC^{1000}$ sessions may be too small when we are talking about a truly synthetic player.

Table 9. Match between *T* and *T'* (winner).

|  | Games Won | | Average # of Moves | |
| --- | --- | --- | --- | --- |
|  | White | Black | White | Black |
| $W_T$ vs. $B_{T'}$ | 384 | <u>616</u> | 451 | <u>159</u> |
| $W_{T'}$ vs. $B_T$ | <u>698</u> | 302 | <u>590</u> | 507 |

Table 10. Match between *P* (winner) and *T*.

|  | Games Won | | Average # of Moves | |
|---|---|---|---|---|
|  | White | Black | White | Black |
| $W_P$ vs. $B_T$ | <u>667</u> | 333 | <u>41</u> | 50 |
| $W_T$ vs. $B_P$ | 580 | <u>420</u> | 517 | <u>498</u> |

Table 11. Match between *P* and *T'* (winner).

|  | Games Won | | Average # of Moves | |
|---|---|---|---|---|
|  | White | Black | White | Black |
| $W_P$ vs. $B_{T'}$ | 837 | <u>163</u> | 60 | <u>213</u> |
| $W_{T'}$ vs. $B_P$ | <u>885</u> | 115 | <u>88</u> | 437 |

We then carried a further set of experiments with a new set of expert players. Therein each white player was moving based on a min-max algorithm[25]. We used five such players, each with an increasing look-ahead value for the min-max algorithm; these values ranged for 1 to 9 (a look-ahead of 2*k*+1 means that we expand *k*+1 moves for the white player and *k* moves for the black one).

As with the pupils and the teachers, we ran a short synthesis tournament on these five min-max trained players and produced $MC_{sum}$. When we compared $MC_{sum}$ to each individual player, we were surprised to see that, in contrast to $T_{sum}$ and $P_{sum}$, $MC_{sum}$ beat its individual min-max origins, in one close case and the rest being comfortable ones. Further confirmation of the quality of $MC_{sum}$ was furnished when we saw that it closely beat $P_{sum}$ (Table 12) and comfortably beat $T_{sum}$ (Table 13), reinforcing our earlier observations about the relative quality of these two synthetic players (as reported in Table 10).

Table 12. Match between *P* and *MC* (winner).

|  | Games Won | | Average # of Moves | |
|---|---|---|---|---|
|  | White | Black | White | Black |
| $W_P$ vs. $B_{MC}$ | 830 | <u>170</u> | 77 | <u>187</u> |
| $W_{MC}$ vs. $B_P$ | <u>989</u> | 8 | <u>11</u> | 227 |

Table 13. Match between *T* and *MC* (winner).

|  | Games Won | | Average # of Moves | |
|---|---|---|---|---|
|  | White | Black | White | Black |
| $W_T$ vs. $B_{MC}$ | 649 | <u>351</u> | 708 | <u>713</u> |
| $W_{MC}$ vs. $B_T$ | <u>986</u> | 14 | <u>28</u> | 511 |

Finally, and this should come as no surprise to the reader (we omit the tables), $MC_{sum}$ outperformed $T'_{sum}$ (but not with the wide margin it beat $T_{sum}$) and $P'_{sum}$ (as generated by a $CC^{10000}$ evolution of $P_{sum}$).

## 5. On the validity and the implication of the results

Capturing a human's playing attitude is, first of all, an exercise in developing adequate infrastructure to codify and store that "attitude". In our case this happens via the reinforcement learning mechanism and the associated approximation of the value function using neural networks.

A couple or a couple of dozens of games, however, only provide for a snapshot of that attitude. Enlarging that snapshot can be accomplished by obtaining more instances of that attitude (i.e. let humans play more games) or by attempting to generalize from the given instances (i.e. attempt to automatically evolve the learned attitude by extensive automatic self-playing), or by combining the two approaches (note the similarity to the cycle of instructive demonstration, generalization and practice trial, as it appears in the robot task learning terminology[11]). Judging what the best combination may be necessarily entails a workflow of human computer interaction activities that may be automated only if we have credible metrics that relate to some notion of game playing quality.

To date we have evaluated learned players by pitting them against each other. An obvious improvement is to also pit them against a benchmark player (for example, a min-max opponent[25], or a scripted opponent). Still the results we have obtained from the tournaments reported in this paper convey the message that "composing immature players results in quality degradation" (excerpts of these games are shown in Table 6 and Table 8). Note that our notion of player composition is based on the co-evolutionary learning taking place during reinforcement learning. Immaturity of the individual players seems to be due to the small number of games involved in CC rounds; when this number is raised from 1,000 to 10,000 (excerpts of these games are shown in Table 9 and Table 11) performance increases. Whether immaturity may be also due to a possibly small duration of the HC sessions (currently 40 games long) is also something that warrants investigation; however, that examination also has to take into account the relative richness (or, the lack of it) of the tactics employed by the human player during the game.[17] When such richness is constrained by the decision of a human player to only pursue a limited number of options, the result is that only a small part of the value function gets a chance to be learned. So, after all, combining small parts of knowledge developed via reinforcement learning is also likely to create confusion since value updates may take less time to affect previously learned parameters.

Learning *and* forgetting what was learned co-exist in a reinforcement learning context; good paths need occasional reinforcement so that learned (useful) value functions do not degrade in approximation quality when alternative paths are explored. Again, however, this seems to be linked to the relative density of high value advice (such is, for example, a concrete playing example by a human), which underlines the conceptual proximity of reinforcement learning to the scaffolding concept in the situated learning approach.[10]

A similar problem exists when a tutor unavoidably displays a behavior that is prone to small errors or deviations, maybe due to physical or mental peculiarities (such is, for example, the case with piloting a helicopter; therein, a best trajectory from a departure point to a destination can be realized as a selection of partially similar real trajectories as followed by a human pilot). In such a context, the learning mechanism has to be designed with a view to subsequently factor out the deviations.[26]

Furthermore, there are interesting implications when one considers the possibility of using tutors who at some point may also act as learning adversaries, attempting to misguide the learning computer. Such unwanted learning may be difficult to mitigate, depending on the strength of the adversarial approach[27]; still, at this point of our research, we view this as an issue of theoretical interest only, since we first need to show how computers can effectively learn from sound advice before tackling adversarial behaviors.

The cautious reader may question the use of a high-school teacher or a student as an expert in our experiments. It is true that these people are not experts but, at the current level of playing *RLGame* as demonstrated by our computers, any reasonable human opponent is expert enough. Moreover, we are keen to promote the interactivity and co-evolutionary learning aspects of our approach and it is known that the pleasure to interact may be a key factor of the success of entertainment robotics (which usually employs less sophisticated machine learning algorithms but quite complex protocols in exploiting the interaction of humans and computers).[9,11]

## 6. Conclusions and future directions

In this paper we have reviewed experimental approaches to developing game players based on input of sample games played by humans and, subsequently, evolved by computer self-play. We have used two player groups, one consisting of high-school students and one consisting of their tutors, for various disciplines. Both groups demonstrated that, while each individual is able to put forward a particular game-playing tactic that can be used as a basis from improved automatic playing, attempting to merge such behaviors in a straightforward fashion does not result in improved automatic game playing. Initially this seems to suggest that we must rethink how to deploy such synthetic approaches to game play learning; it looks like composition which can exploit parallelism is not as easy as intuition would have lead us believe.

Interactive evolution is a promising direction. In such a course, one would ideally switch from focused human training to autonomous crawling between promising alternatives. But, as we have discovered during the preparation of this work, the interactivity requirements of the process of improving the computer player is very tightly linked to the availability of a computer-automated environment that supports this development. It is a must to strive to put the expert in the loop as efficiently as possible.

In terms of the experiments described above, we have noticed several features of an experimentation system that we have deemed indispensable if one views the project from the point of system efficiency. Such features range from being able to easily design an experimentation batch, direct its results to a specially designed database (to also facilitate reproducibility), automatically process the game statistics and observe correlations, link experimentation batches in terms of succession, while at the same time being able to pre-design a whole series of linked experiments with varying parameters of duration and succession and then guide the human player to play a game according to that design. We needs to improve interaction if we aim for interactive evolution

In the near future we aim to deploy at a larger scale the web-based experimentation engine to collect input from more human players and to link the self-playing sessions to grid-enabled computing for improved efficiency and scalability. In doing so we aim to focus human effort where it is mostly needed (i.e. to provide concrete examples of plat according to an un-specified strategy) and integrate seamlessly with machine intelligence.

**Acknowledgments**